%% file: main.tex
\documentclass[pdflatex,sn-mathphys-num]{sn-jnl}
\usepackage{graphicx}
\usepackage{amsmath,amssymb}
\usepackage{booktabs,tabularx,array}
\usepackage{placeins}
\hypersetup{pdftitle={What Do We Expect from LLMs? Mapping the Design of LLM Benchmarks},pdfauthor={Chao Wang}}
\newcommand{\datasetN}{14,767}
\begin{document}
\title[What Do We Expect from LLMs?]{What Do We Expect from LLMs? Mapping the Design of LLM Benchmarks}
\author[1]{\fnm{Chao} \sur{Wang}}
\affil[1]{\orgname{Independent Researcher}}
\abstract{Benchmarks are central to how progress in large language models (LLMs) is assessed and communicated. Yet model rankings alone reveal little about how evaluation requirements themselves are changing. The expanding variety of benchmarks offers another perspective: what researchers expect LLMs to do, and what they count as successful performance. We systematically map \datasetN\ papers introducing or updating evaluation resources from arXiv submissions between January 2022 and August 2026. Using staged screening and automated full-text coding, we examine changes in target systems and domains, evaluation materials and conditions, and scoring mechanisms. The collection shows growing emphasis on action, interaction, and professional applications, while established and newer design elements frequently coexist. Model participation also develops unevenly: LLM-based scoring grows within both agent and non-agent groups, whereas model-generated materials show no comparable sustained increase in recent cohorts. These findings illuminate how public research translates capability expectations into concrete tests and criteria for success. As AI participates in constructing tests, performing tasks, and judging responses, they also raise a question: does expanding evaluation provide more independent evidence, or risk reproducing the preferences and blind spots of its participating models?}
\keywords{large language models, benchmarks, evaluation design, systematic mapping, metaresearch, research expectations}
\maketitle

\section{Introduction}\label{sec:intro}
Benchmarks provide much of the public evidence used to compare large language models (LLMs) and assess claims of progress. Their importance extends beyond ranking: which tasks enter a benchmark, and which outcomes earn credit, help define what it means for a model to improve. As LLMs are proposed for more uses, benchmark development has expanded across capabilities, professional domains, and forms of interaction \cite{chang2023survey,ni2025benchmarks}.

This expanding collection offers a way to study the demands being placed on LLMs, not just how well models meet them. A benchmark for medical questions, software development, or tool-mediated tasks turns a proposed use into concrete requirements: what inputs a model receives, what it must produce or do, and how success is recognized. Benchmark choices encode judgments about what matters \cite{dehghani2021lottery}. We therefore study benchmark designs as \emph{research expression}: the translation of capability expectations into testable requirements.

This perspective calls for a broader collection than a study of widely used benchmarks. A resource can document a proposal about what an LLM should be able to do even if it rarely appears in later experiments. Publication leaves a record of the tasks, conditions, and assessment procedures its designers chose. Its value as evidence of a research proposal is therefore distinct from its influence as a measurement tool.

We examine the systems and domains these resources target, the materials and conditions they use, and the ways they score performance. Tracking these choices reveals how proposed test requirements change over time. They give only a partial view of expectations: they neither recover every fine-grained capability nor directly measure authors' beliefs, industrial demand, or community consensus.

Three questions guide the mapping:
\begin{enumerate}
\item \textbf{RQ1:} Which system types and professional or knowledge domains are represented in proposed LLM evaluation resources, and how does their distribution vary over time?
\item \textbf{RQ2:} Through which materials, evaluation setups, modalities, and task languages are these requirements operationalized?
\item \textbf{RQ3:} Which scoring mechanisms assess performance, and how do their distributions and combinations with target types and material origins vary?
\end{enumerate}

Our mapping of \datasetN\ arXiv paper records connects three changes that are often considered separately: the systems and domains being tested, the conditions of testing, and the sources of judgments about success. It distinguishes expanding task requirements from changes in how evaluation is carried out, and greater model participation from a uniform move toward automation. This perspective also raises a question about evaluation evidence itself: when models help construct tests and judge responses, how are their assessments checked against the requirements those tests are intended to represent?

\section{Methods}\label{sec:methods}
\subsection{Design, scope, and unit of analysis}
We conduct a systematic literature mapping with descriptive and exploratory analyses. The frame covers arXiv papers first submitted from 1 January 2022 through 31 August 2026. Eligibility is assessed from the exact version supplied to the coding model. No citation, popularity, or adoption weights are applied: the study describes identified resource proposals and their designs, not their influence.

The statistical unit is a paper record. Each arXiv paper contributes one retained exact-version record and is not counted repeatedly across subject categories. Multiple formal evaluation modes within a paper are represented by multiple labels. We do not perform exhaustive cross-paper benchmark-family resolution. Consequently, \datasetN\ paper records are not \datasetN\ independent benchmark families. Paper-level aggregation also does not identify which labels belong to the same component of a multi-resource paper.

\subsection{Eligibility}
A paper must introduce or substantially update a reusable evaluation resource, specifying tasks for evaluated systems, evaluation materials or a task-generation mechanism, and a way to score performance. Eligible instruments may include evaluation datasets, test suites, interactive environments, task generators, or operational evaluation protocols. The resource must be presented as a contribution that later researchers could use to evaluate new systems, rather than merely as internal experimental material. A benchmark may be released alongside a new model or method; it need not be the paper's sole or primary contribution.

Using existing benchmarks, releasing training data alone, or assembling internal held-out tests, ablations, or case studies without an independent resource contribution is insufficient. A new scoring, auditing, or analytical method alone is also insufficient unless accompanied by new or substantially updated test materials, tasks, environments, or generators. The word ``benchmark,'' code availability, and multiple baselines are not individually sufficient evidence. We assess the resource described by the paper; we do not independently verify that every external dataset or service remains downloadable or executable.

Target scope is restricted to generative text LLMs, generative multimodal systems centered on language models, and agents or embodied systems centered on such models. The target is the system formally assigned scores or rankings. Using an LLM only to generate data or judge another algorithm does not qualify. Resources with conventional classifiers, retrievers, detectors, or other non-LLM systems as substantive formal co-targets are excluded; explicitly designated controls, ceilings, or structural reference baselines do not automatically trigger exclusion. Unresolved resource or target eligibility is not included in the main analysis.

\subsection{Screening and full-text coding}
We combine arXiv OAI-PMH metadata with frozen monthly sources and assign study months by first submission. A local classifier (M0) ranks each month's papers using title, abstract, and category features and routes approximately the highest-scoring 15\%, including ties, to abstract screening. Categories are not hard exclusion rules. M0 was trained under an earlier, stricter main-contribution definition; final eligibility permits resource and method co-release.

GLM-5.3-Flash screens routed abstracts for a clearly introduced reusable evaluation resource and an LLM target. Only records clearly satisfying both conditions proceed; uncertain records are not routed. A final full-text request reassesses eligibility and assigns seven coding fields. Abstract acceptance does not force inclusion. This selective pipeline may miss eligible papers before full-text assessment.

The model receives text extracted from exact-version PDFs, including body text, appendices, and references without deliberate section cropping. Unavailable or unprocessable sources do not enter the final analysis and remain accounted for in the selection flow. Oversized inputs are not silently truncated. Both model stages use the provider identifier \texttt{glm-5.3-flash}; prompts, schemas, source versions, and outputs are archived. Appendix~\ref{app:technical} gives acquisition, extraction, and request details.

\subsection{Coding fields}

\begin{table}[htbp]
\caption{Seven coding fields and their operational boundaries. A paper may receive multiple labels within each field.}\label{tab:fields}
\small\begin{tabularx}{\textwidth}{@{}p{0.22\textwidth}X@{}}\toprule
Field & What is recorded \\\midrule
Target system & Answering models submit answers or generated artifacts; agents take tool or environment actions; embodied systems control bodies. Executing submitted code for scoring does not by itself make the target an agent.\\
Evaluation domain & Knowledge or application domains actually tested, not architecture, modality, or incidental story setting. General/cross-domain and specific domains may coexist.\\
Evaluation setup & Fixed items/tasks/starting states, runtime-generated scored items, and interaction with action-dependent feedback. Fixed tasks can require interaction.\\
Modality & Formal inputs and scored outputs: text, image, video, audio, code, and structured data; not storage format or paper illustrations.\\
Material origin & Sources of evaluation content: human-written or recorded real-world material, generative models, and programs or simulations. Human design or review alone does not establish human-authored content.\\
Scoring source & Reference/metric scoring, execution/environment outcomes, human judges, LLM judges, or other learned scorers. These identify what produces an item-level score, not how scores are averaged.\\
Task language & Natural languages in actual task inputs, instructions, or required responses; not paper language, display-only translation, or programming syntax.\\\bottomrule
\end{tabularx}
\end{table}

Table~\ref{tab:fields} summarizes the fields. Labels describe the introduced resource, rather than unrelated training or ablation experiments.

\begin{samepage}
Where permitted by the field, \texttt{unclear} records an unresolved classification and \texttt{not\_reported} records missing source information; neither means that a property is absent. The domain label \texttt{other} denotes a specified domain outside the listed categories, not missing information.
\end{samepage}

\subsection{Quality checks and validation scope}
Programs check identifiers, joins, flow counts, field validity, and denominators; a separate CSV projection checks key calculations, and archived inputs reproduce the deterministic analysis. Here, a valid field passes the output-format and allowed-label checks; validity does not establish that its labels are correct. Invalid fields are treated as missing without discarding the paper's other valid fields. Records with unresolved eligibility are excluded.

Development combined model trials, consistency comparisons, and source checks to clarify eligibility and coding rules. In a subsequent year-stratified random sample of 100 final included records, author review and source-based clarification confirmed 99 as eligible. Supplementary checks examined three upstream nonretention pools. These checks assess eligibility, not the accuracy of the seven design fields; sampling and results are summarized in Appendix~\ref{app:validation}.

\subsection{Analysis}
We report paper counts and label shares. Each field uses its valid-record denominator; cross-field comparisons require all relevant fields to be valid. Multi-label percentages and overlapping domains need not sum to 100\%. Permitted special states remain in the principal denominators, with additional summaries excluding them. An unselected label does not establish absence.

We compare complete years in 2022--2025 and matched January--August cohorts in 2022--2026. Detailed comparisons use the larger 2024--2026 matched cohorts. Small groups, particularly those with fewer than 20 records, are not emphasized as trend evidence. Cohorts use first-submission dates, whereas labels describe the version read, not necessarily the original release design.

To distinguish changes in group composition from changes within groups, we use Kitagawa's symmetric decomposition \cite{kitagawa1955components}, primarily grouping agent and non-agent records. We also examine generated-material/LLM-scoring co-occurrence and sensitivity to alternative windows, version-year subsets, and assumed errors. The decompositions are descriptive, not causal; error budgets are hypothetical, not measured. Formulas and supporting comparisons appear in Appendix~\ref{app:supplement}. Analyses are exploratory and were developed after inspecting data, without selection by statistical significance.

\section{Results}\label{sec:results}
\subsection{Collection and temporal coverage}
\begin{figure}[htbp]\centering
\includegraphics[width=\textwidth]{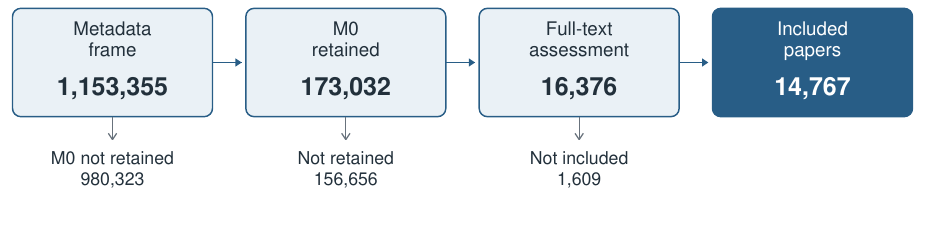}
\caption{Screening and final analysis flow for January 2022--August 2026, with one exact-version record per paper. Detailed reasons for non-retention and the complete count breakdown are reported in Appendix~\ref{app:availability}. Box sizes do not encode counts.}\label{fig:flow}
\end{figure}

From a metadata frame of 1,153,355 papers, the staged pipeline identifies \datasetN\ included records; 14,736 have valid coding for all seven fields. Figure~\ref{fig:flow} accounts for screening and exclusions, and Table~\ref{tab:years} shows annual coverage. Field-specific missingness and special states are detailed in Appendix~\ref{app:availability}.

\begin{table}[htbp]\centering
\caption{Included records by first-submission year. The incomplete 2026 year is not directly compared with full earlier years as an annual growth rate.}\label{tab:years}
\begin{tabular}{lrr}\toprule
Year & Observed months & Included records\\\midrule
2022 & January--December &160\\
2023 & January--December &861\\
2024 & January--December &2,735\\
2025 & January--December &5,277\\
2026 & January--August &5,734\\\bottomrule
\end{tabular}
\end{table}

Agent/tool-system and LLM-scoring shares rise in both temporal series. Interaction is more frequent in later cohorts, although the small early matched cohorts do not form a monotonic sequence. Fixed collections remain common. Generated materials expand earlier, then fluctuate around one half in recent cohorts (Appendix~\ref{app:windows}).

The following sections examine the larger 2024--2026 January--August cohorts. Unless another window is stated, two-endpoint comparisons refer to 2024 and 2026; three-value sequences include 2025 between them.

\subsection{Target systems and professional domains}
\begin{figure}[htbp]\centering
\includegraphics[width=\textwidth,height=.62\textheight,keepaspectratio]{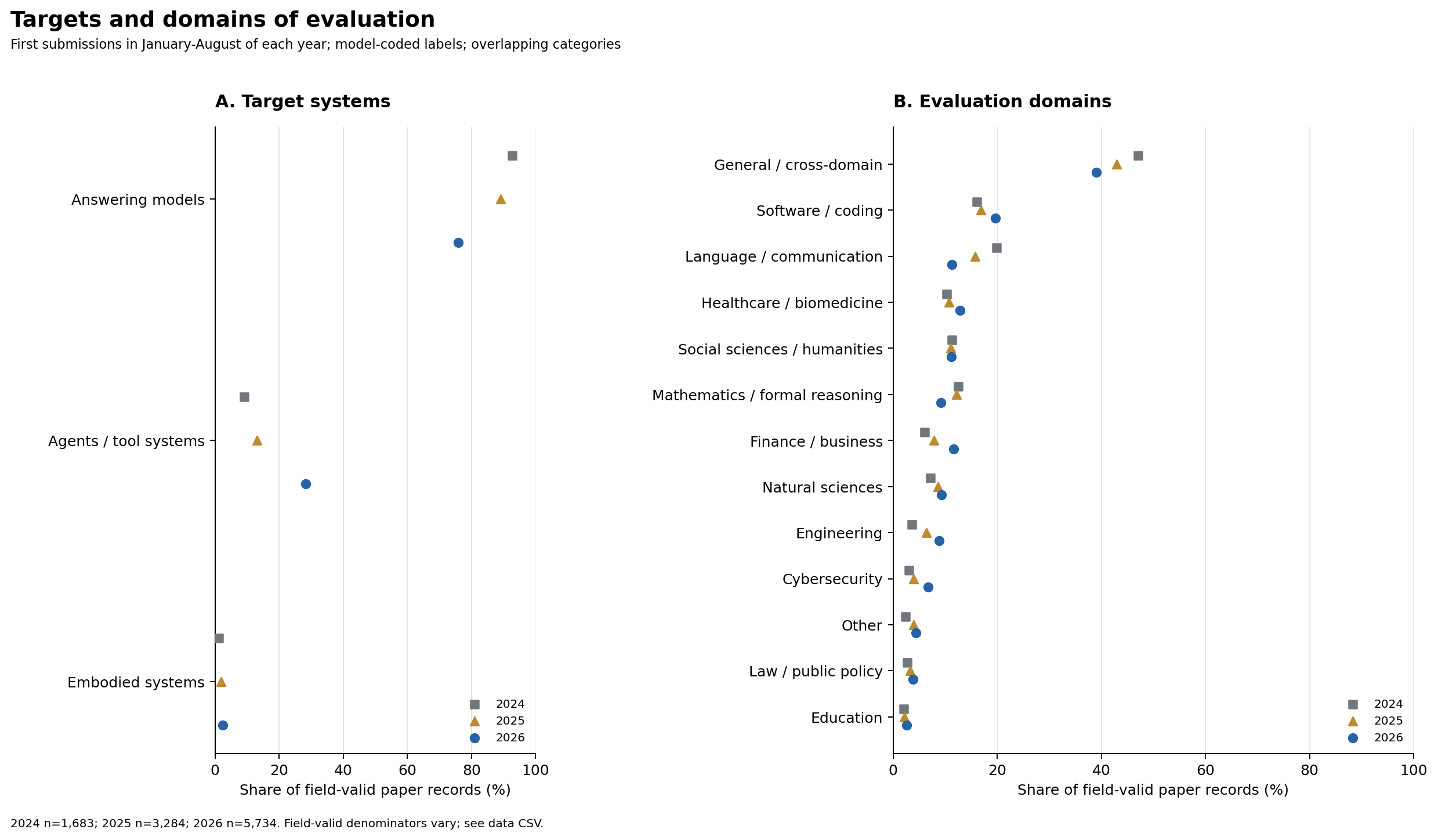}
\caption{Target systems and all observed domain categories in January--August 2024, 2025, and 2026. Labels overlap.}\label{fig:targets}
\end{figure}

The agent/tool-system share rises from 152/1,683 (9.0\%) in January--August 2024 to 428/3,284 (13.0\%) in 2025 and 1,624/5,734 (28.3\%) in 2026. Answering models remain common: the share of papers with answer or artifact-submission tasks falls from 92.8\% in 2024 to 75.8\% in 2026. Over the same interval, the embodied-system share increases from 1.1\% to 2.3\%. These are overlapping labels, not mutually exclusive system categories (Fig.~\ref{fig:targets}).

In healthcare and biomedicine, matched counts rise from 174 in 2024 to 353 in 2025 and 736 in 2026, while shares rise from 10.3\% to 10.7\% and 12.8\%. The small initial share change therefore accompanies approximately a doubling of papers. Finance/business, engineering, and cybersecurity also grow in both counts and shares (Table~\ref{tab:domains-matched}; Fig.~\ref{fig:targets}). Full-year domain counts are retained in Appendix~\ref{app:windows}.

Falling shares do not necessarily mean fewer proposals. Language/communication records increase from 334 to 647 and mathematics/formal reasoning from 210 to 528, despite declining shares. Nor is specialization a uniform shift: general/cross-domain coding first rises and then falls over the longer window, and can coexist with specific domains. Detailed trajectories are retained in Appendix~\ref{app:windows}.

\input{domain_matched}

\subsection{Evaluation conditions and materials}
Interactive evaluation grows alongside fixed task collections. Its share rises from 160/1,683 (9.5\%) in January--August 2024 to 1,615/5,733 (28.2\%) in 2026. Fixed item, task, or starting-state collections remain at approximately 95\% (95.7\% and 95.0\%). Both labels co-occur in 132/1,683 (7.8\%) and 1,441/5,733 (25.1\%) records, respectively. Fixed collections have therefore not simply been displaced by interactive evaluation. Runtime-generated scored items change less over the same interval, from 6.7\% to 7.5\%.

Text remains nearly universal, with structured data and audio appearing more frequently in recent cohorts. English-only coding remains common but shows no consistent annual increase across both temporal series. Here it means only English was identified, not that all other languages were ruled out. Detailed modality and language sequences, including non-monotonic changes and coding limitations, appear in Appendix~\ref{app:materials}.

\begin{figure}[htbp]\centering
\includegraphics[width=\textwidth,height=.62\textheight,keepaspectratio]{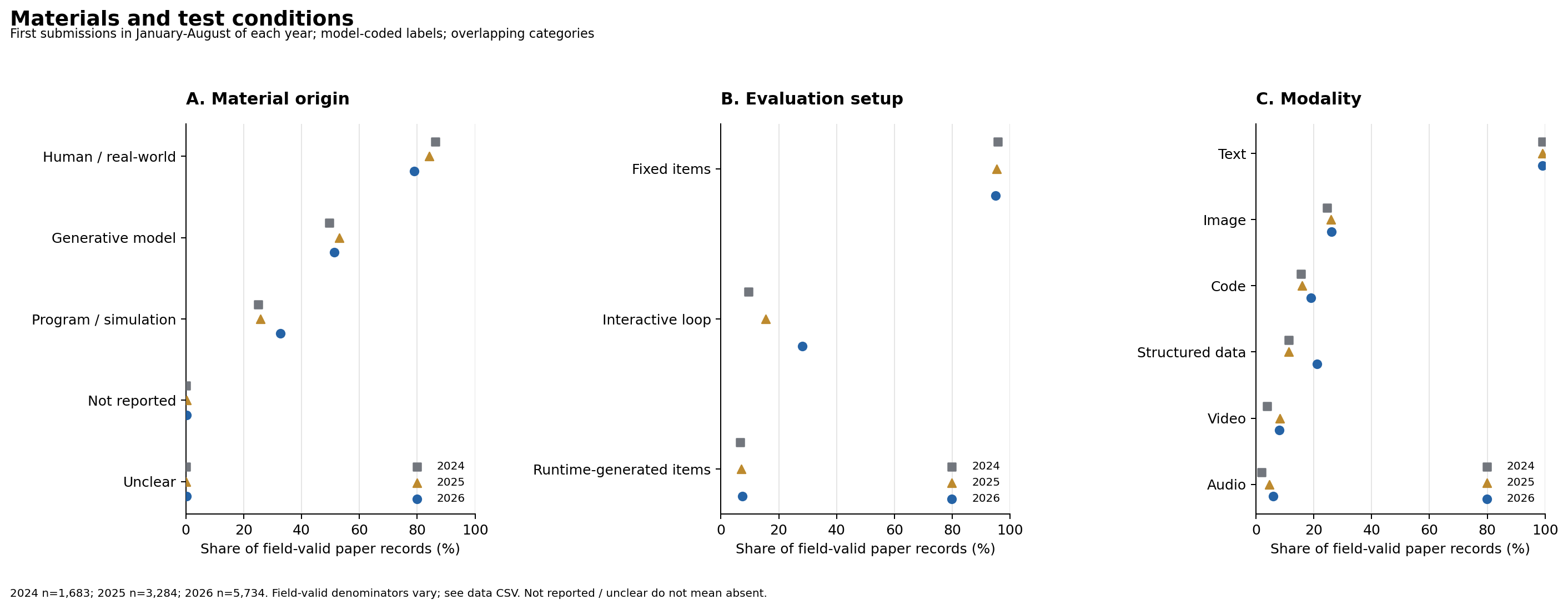}
\caption{Material origins, evaluation setups, and modalities in January--August 2024, 2025, and 2026. Program/simulation and generative-model material are separate categories; missing or unclear coding does not establish absence.}\label{fig:materials}
\end{figure}

Generative-model material appears in 49.5\%, 53.1\%, and 51.3\% of the 2024--2026 matched cohorts. This recent plateau follows an earlier expansion, from 15.6\% of full-year 2022 records to 49.5\% in 2024. Program/simulation material grows, while human/real-world material remains common; sources often coexist (Fig.~\ref{fig:materials}).

\subsection{Scoring changes within target-system groups}
\begin{figure}[htbp]\centering
\includegraphics[width=\textwidth,height=.62\textheight,keepaspectratio]{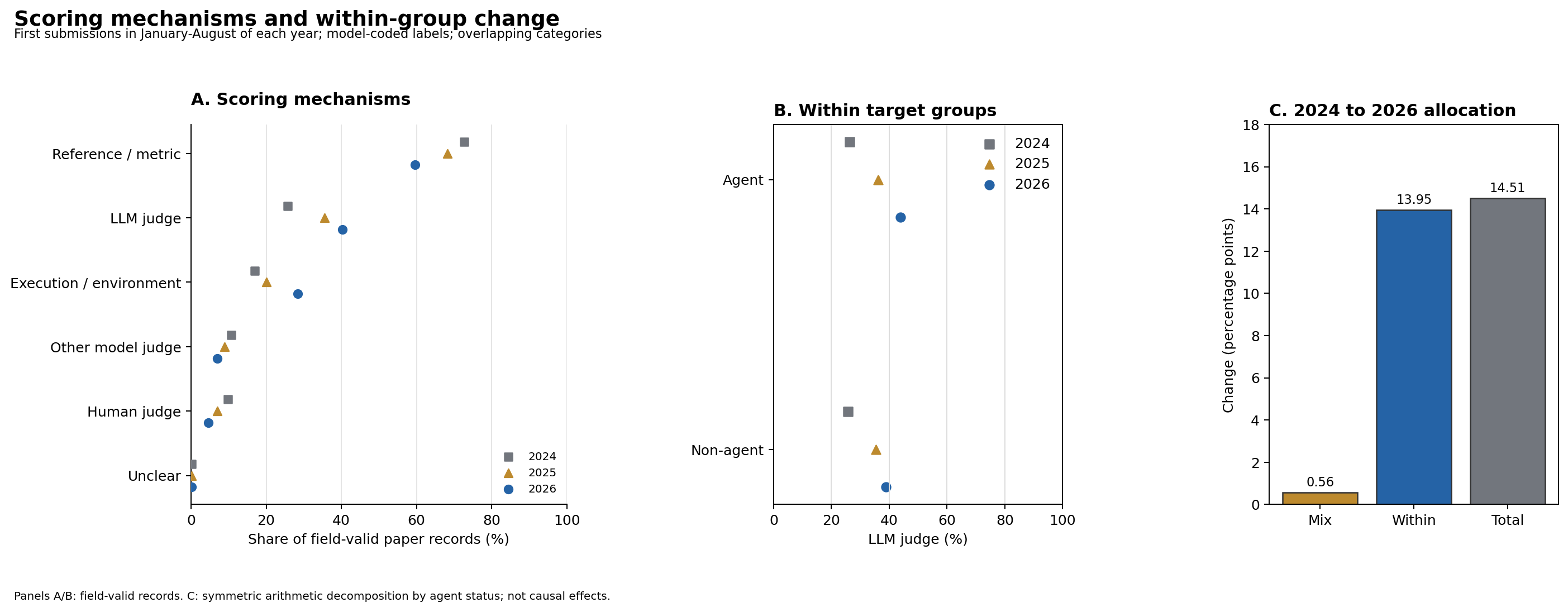}
\caption{Scoring mechanisms and LLM scoring within agent and non-agent groups in January--August 2024--2026, with symmetric decomposition of the 2024--2026 aggregate change. Decomposition values are percentage points and are not causal effects.}\label{fig:scoring}
\end{figure}

LLM scoring becomes more common both overall and within target-system groups (Fig.~\ref{fig:scoring}). Its overall share increases from 433/1,681 (25.8\%) in January--August 2024 to 1,166/3,283 (35.5\%) in 2025 and 2,309/5,734 (40.3\%) in 2026. Between 2024 and 2026, execution/environment scoring rises from 16.9\% to 28.4\%; reference/metric scoring changes from 72.6\% to 59.6\%, and direct human judging from 9.8\% to 4.7\%. Scoring mechanisms can coexist, so these movements do not establish replacement or a change in scoring quality.

Between the 2024 and 2026 matched cohorts, LLM scoring rises from 40/152 (26.3\%) to 714/1,624 (44.0\%) among agent records, and from 393/1,529 (25.7\%) to 1,595/4,110 (38.8\%) among non-agent records. The growing agent share alone therefore cannot account for the overall increase. Under this binary grouping, the aggregate 14.51-percentage-point change comprises 0.56 points from changing group proportions and 13.95 points from changes within groups.

Alternative time windows, same-year-version subsets, and finer target-system partitions also show a large within-group contribution. The pattern is not universal across domains: engineering's LLM-scoring share decreases slightly, while the other 12 observed domain groups increase. Exact comparisons are in Appendix~\ref{app:scoring}.

\subsection{Generated materials and LLM scoring often coexist with other mechanisms}
\begin{figure}[htbp]\centering
\includegraphics[width=\textwidth,height=.62\textheight,keepaspectratio]{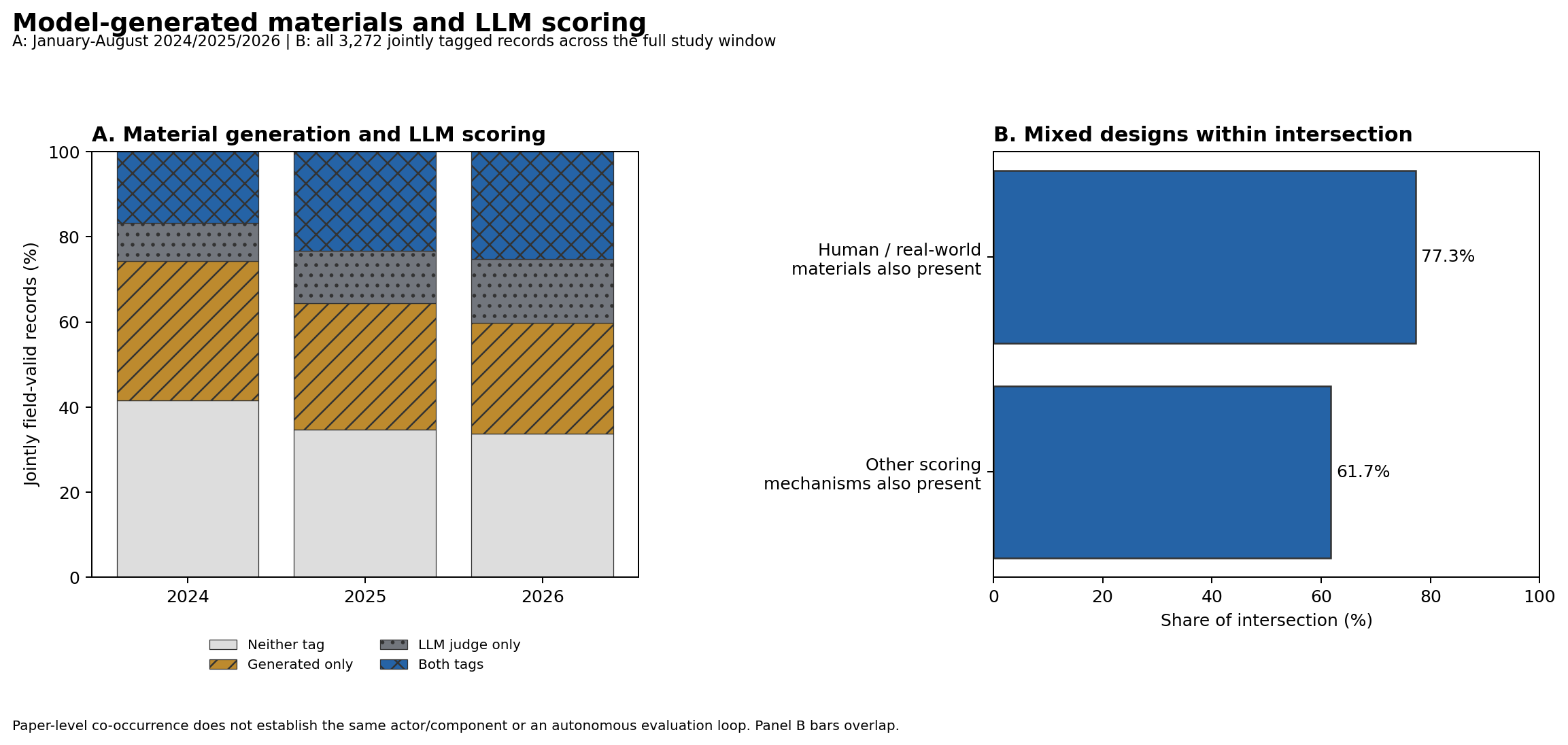}
\caption{Generation--scoring combinations. Left: mutually exclusive combinations in January--August 2024, 2025, and 2026. Right: coexistence of additional mechanisms among the 3,272 records containing both generated material and LLM scoring across January 2022--August 2026. The conditional denominator on the right differs from the cohort denominators on the left.}\label{fig:joint}
\end{figure}

Across the full January 2022--August 2026 window, 3,272/14,763 records with both fields valid (22.2\%) contain generative-model material and LLM scoring. Of these, 2,530 (77.3\%) also contain human/real-world material and 2,020 (61.7\%) another scoring mechanism. These conditional groups overlap. The combination is better described as mixed design than as fully autonomous evaluation; mixture alone establishes neither supervision nor validity (Fig.~\ref{fig:joint}).

The joint share rises from 16.8\% in the 2024 matched cohort to 25.3\% in 2026. This does not imply tighter coupling: co-occurrence relative to the product of marginal shares decreases (Appendix~\ref{app:association}). Paper-level labels do not identify the same model, component, or item across roles and therefore do not establish a self-evaluation loop.

\subsection{Sensitivity summary}
The four larger increases---agent systems, interaction, LLM scoring, and execution/environment scoring---remain positive under an assumed maximum 5\% binary-label error in each endpoint cohort. Smaller material, fixed-collection, and English-only differences can change direction under the same scenario. These conditional bounds support emphasizing larger changes, but do not establish that actual errors meet the assumed budgets or address upstream omissions. Full bounds, a separate ineligible-record removal scenario, and the limited effect of two documented interaction omissions appear in Appendix~\ref{app:sensitivity}.

\FloatBarrier
\section{Discussion}\label{sec:discussion}
\subsection{More proposals ask models to act and interact}

Fixed task banks can support interactive evaluation. A predefined task bank can allow feedback and action during execution while retaining comparable initial conditions. The co-occurrence of these labels cautions against treating ``static'' and ``interactive'' as successive eras.

For example, Asuka-Bench fixes 50 web-development tasks but lets a code agent revise its implementation after browser-based evaluation and feedback \cite{wang2026asuka}. The task bank is fixed; the process of completing a task is interactive.

\subsection{Professional-domain growth in counts and shares}
Growth in professional-domain proposals makes application-specific demands and failure modes visible. Counts and shares together distinguish an expanding area from one that merely gains prominence relative to others. This records research attention, including attempts to expose weaknesses, rather than adoption or endorsement of current models.

\subsection{Model use changes differently in materials and scoring}
The different trajectories of generated materials and LLM scoring argue against combining model participation into a single measure of automation. Models can play distinct roles in constructing evaluation content and determining whether a response succeeds.

WebArena, for example, programmatically checks website states for some tasks and uses GPT-4 to judge semantic equivalence to reference answers for others \cite{zhou2023webarena}. Multiple scoring labels can therefore describe different tasks within one benchmark, rather than multiple judges assessing every response.

Asuka-Bench illustrates all three roles: an LLM helps construct task specifications from source queries, LLM-based code agents implement them, and an LLM-based evaluator judges browser-observed behavior \cite{wang2026asuka}. Model participation in construction, performance, and assessment does not by itself mean human-free benchmark creation.

Role overlap makes it important to distinguish evaluation scale from independent evidence. Beyond documented LLM-judging biases \cite{zheng2023judge}, experiments by Li et al. show that judges can favor student models trained on synthetic data from related generators \cite{li2025preference}. That training-data setting differs from benchmark construction, but it demonstrates why nominally separate model roles need not provide independent judgments. Larger evaluation pipelines could reproduce shared preferences rather than subject them to additional scrutiny.

A broader concern is which demands become measurable. Messeri and Crockett warn that widespread reliance on AI tools may narrow scientific questions, methods, and viewpoints \cite{messeri2024illusions}. Applied to benchmarks, this suggests a conditional risk: tasks that models can readily generate and judge may receive disproportionate attention. If the resulting scores guide model selection and optimization, preferred behaviors could be reinforced. Our mapping does not establish these feedback processes. It instead motivates checking whether model-mediated tests remain anchored to their intended requirements through validated execution outcomes, external task results, or domain expertise.

\section{Limitations and threats to validity}\label{sec:limits}
\subsection{Coverage and coding uncertainty}
The collection covers identified arXiv proposals, not all public or private evaluation resources. Monthly routing, the earlier M0 training target, and selective abstract screening may miss early terminology or co-released resources disproportionately. Final full-text screening cannot recover those omissions, so differences across years or domains may partly reflect selection.

The eligibility audits provide sample-based quality checks, but do not establish end-to-end recall or seven-field coding accuracy. Known coding issues include interaction omissions and material-origin, scoring, and language errors (Appendix~\ref{app:validation}); their prevalence and distribution across years and domains remain uncertain. Sensitivity analyses show which directions survive assumed error budgets, not whether actual errors meet those budgets; they do not address upstream omissions or group misclassification.

\subsection{Unit, timing, and interpretation}
Paper records are not independent benchmark families, and paper-level co-occurrence cannot identify which components share a design. Labels describe the version read, which may include later revisions, while cohort dates refer to first submission. Same-year-version checks cannot reconstruct original release designs. PDF extraction may also lose information from tables, formulas, or layouts.

The seven fields describe broad design choices rather than fine-grained abilities, adoption, or construct validity \cite{jacobs2021measurement,bean2025validity}. Ambiguous domain boundaries require care when interpreting small differences. Our exploratory decompositions are descriptive rather than causal, and \emph{research expectations} refers to documented test requirements, not measured beliefs or community consensus.

\section{Conclusion}
The expansion of LLM benchmarks records more than a growing supply of tests. It documents how researchers turn proposed uses into task demands, interaction conditions, and criteria for success. In this collection, broader professional applications and greater emphasis on action coexist with established task structures rather than replacing them. The different trajectories of model-generated materials and model judging further show that evaluation is changing along distinct dimensions, not simply becoming more automated.

As AI participates in making tests as well as taking and judging them, an important question remains: can a larger evaluation system still expose limitations that its participating models may share? This mapping makes those design choices visible and comparable, providing a basis for investigating their independence and effectiveness.

\backmatter
\section*{Data and code availability}
The dataset, frozen coding prompts and schemas, and analysis code are publicly available at \url{https://github.com/xxcg322/LLM-Bench-Map}. Tables and figures can be rebuilt from the archived outputs without new model calls.

\section*{Declaration of generative AI and AI-assisted technologies in the manuscript preparation process}
During the preparation of this work, the author used generative AI tools to assist with code development and to improve the readability and language of the manuscript. After using these tools, the author reviewed and edited the content as needed and takes full responsibility for the content of the published article.

\begin{appendices}
\renewcommand{\theHtable}{appendix.\Alph{section}.\arabic{table}}
\input{technical_details}
\FloatBarrier
\input{supporting_results}
\FloatBarrier
\section{Eligibility audit and coding checks}\label{app:validation}
\subsection{Author eligibility review}
After freezing the production outputs, we drew random samples from the final included collection and three disjoint upstream nonretention pools (Table~\ref{tab:eligibility-audit}). Sampling used sorted identifiers, fixed random seeds, and selection without replacement within the specified strata. Titles and abstracts were retained unchanged, and records were presented in shuffled order.

The author assessed eligibility using titles and abstracts, consulting full texts where needed. Borderline judgments were clarified against the sources before finalization. The sampling purpose was known, but individual production labels and model explanations were not displayed. The review covered eligibility only, not the seven design fields.

\begin{table}[htbp]
\centering
\caption{Final author eligibility judgments in four separately sampled pools. Counts are observed sample results, not pooled estimates of population error.}\label{tab:eligibility-audit}
\begin{tabular}{lrrr}
\toprule
Sampling pool & Population & Reviewed & Judged eligible\\
\midrule
M0 non-retained & 980,323 & 200 & 0\\
Abstract undetermined & 25,079 & 100 & 1\\
Abstract explicitly excluded & 131,525 & 50 & 0\\
Final included & 14,767 & 100 & 99\\
\bottomrule
\end{tabular}
\end{table}

The M0 and abstract-undetermined samples contained 40 and 20 records per year, respectively, with quotas distributed as evenly as possible across months before random selection within each month. The explicitly excluded sample contained 10 random records per year. For final inclusions, approximately proportional allocation yielded 1, 6, 18, 36, and 39 records from 2022--2026; the 2026 frame ends in August. Stratum sizes and selection probabilities are retained in the audit records.

No final judgment remained undetermined. The one eligible record found upstream was confirmed by full-text review. These observations support a bounded eligibility check, not proof of negligible omissions: the large M0 pool is sparsely sampled, other pipeline exits are not covered, and the samples do not support precise year-specific error estimates. Population estimates would require the recorded sampling weights. The reported analyses retain the frozen production cohort.

\subsection{Coding checks}
Development and targeted source checks identified interaction omissions and material-origin, scoring, and language-coding errors. These diagnostic records are retained separately from the author eligibility audit. They document limitations of automated coding but do not provide a representative estimate of seven-field accuracy.

\end{appendices}

\setlength{\bibsep}{3pt}
\bibliography{references}
\end{document}

%% file: domain_matched.tex
\begin{table}[tb]\centering\small
\caption{Professional and application domains: January--August in each year, 2022--2026. Cells show paper counts with within-year percentages in parentheses. Domain coding is valid for all included records; categories overlap. Counts are identified papers, not independent families or adoption measures.}\label{tab:domains-matched}
\setlength{\tabcolsep}{3pt}
\begin{tabular}{lrrrrr}\toprule
Domain & 2022 & 2023 & 2024 & 2025 & 2026\\\midrule
Healthcare/biomedicine & 5 (7.2) & 28 (7.4) & 174 (10.3) & 353 (10.7) & 736 (12.8)\\
Finance/business & 4 (5.8) & 23 (6.1) & 102 (6.1) & 257 (7.8) & 667 (11.6)\\
Engineering & 1 (1.4) & 19 (5.1) & 61 (3.6) & 210 (6.4) & 509 (8.9)\\
Cybersecurity & 1 (1.4) & 9 (2.4) & 51 (3.0) & 130 (4.0) & 388 (6.8)\\
Software/coding & 13 (18.8) & 70 (18.6) & 270 (16.0) & 555 (16.9) & 1,128 (19.7)\\
Natural sciences & 3 (4.3) & 29 (7.7) & 121 (7.2) & 283 (8.6) & 535 (9.3)\\
Law/public policy & 2 (2.9) & 12 (3.2) & 46 (2.7) & 107 (3.3) & 219 (3.8)\\
Education & 0 (0.0) & 9 (2.4) & 35 (2.1) & 71 (2.2) & 149 (2.6)\\
Social sci./humanities & 8 (11.6) & 53 (14.1) & 191 (11.3) & 364 (11.1) & 640 (11.2)\\
Language/communication & 39 (56.5) & 123 (32.7) & 334 (19.8) & 517 (15.7) & 647 (11.3)\\
Math./formal reasoning & 5 (7.2) & 54 (14.4) & 210 (12.5) & 401 (12.2) & 528 (9.2)\\
General/cross-domain & 19 (27.5) & 168 (44.7) & 792 (47.1) & 1,412 (43.0) & 2,237 (39.0)\\
Other & 1 (1.4) & 7 (1.9) & 41 (2.4) & 129 (3.9) & 254 (4.4)\\
\midrule
All included papers & 69 & 376 & 1,683 & 3,284 & 5,734\\\bottomrule
\end{tabular}
\end{table}

%% file: technical_details.tex
\section{Source preparation and operational details}\label{app:technical}
\subsection{Metadata frame and staged screening}
The frame combines arXiv OAI-PMH harvesting with frozen monthly sources. The newly harvested historical snapshot extends to 7 September 2026 in metadata modification time; study months are assigned locally from first-version submission dates, not OAI modification dates. Subject categories are not hard exclusion rules. Source manifests retain identifiers, versions, month assignments, and hashes, and matching existing sources are reused.

The first stage, M0, ranks papers within each month using title, abstract, and category features. The frozen classifier uses character TF-IDF with weighted logistic SGD. Approximately the highest-scoring 15\% are retained, including ties at the boundary. Scores are routing values, not calibrated eligibility probabilities. M0 was developed against an earlier, stricter main-contribution definition, whereas final eligibility allows resource and method co-release. That history can affect recall and is not corrected by later full-text screening.

In the second stage, GLM-5.3-Flash reads each routed title and abstract and returns two controlled decisions: whether a concrete reusable evaluation resource is clearly introduced, and whether formal targets fall within LLM scope. Only records clearly satisfying both conditions proceed. Any uncertain decision is not routed. This intentionally selective rule can miss papers whose full text contains information absent from their abstract.

The final stage independently reassesses resource and target eligibility and assigns seven fields in one full-text request. Passing abstract screening does not force inclusion. The model returns controlled JSON. Document-level provenance, prompts, schemas, and request receipts are retained separately.

\subsection{Full-text preparation and coding}
Exact-version PDFs are reused where available and downloaded otherwise. Unavailable sources are removed from the effective full-text set but remain in flow accounting. Text extraction uses PyMuPDF with a pypdf fallback for exceptions, accompanied by document identity, page continuity, text availability, and hash checks. These are technical checks, not manual verification that every table, formula, or reading order has been preserved. Models receive extracted text rather than page images.

The final input retains body text, appendices, and references without deliberate section cropping. Inputs exceeding the protective size limit are not silently truncated. There is normally one qualification-and-label request per paper; matching valid outputs are reused and limited confirmed technical failures are handled by bounded recovery. Repeated semantic sampling is not used to select favorable answers.

Both abstract and final coding use the provider identifier \texttt{glm-5.3-flash}. Prompt, schema, and request-code hashes are frozen; final request settings are listed below. The API identifier does not freeze provider weights or guarantee identical future responses. Deterministic reproducibility concerns analysis of archived outputs.

\subsection{Coding request settings}
Final requests specify high reasoning effort, disabled sampling, structured JSON, a 3,200-token output limit, and a 1,000,000-byte UTF-8 input guard.

%% file: supporting_results.tex
\section{Supporting comparisons and sensitivity details}\label{app:supplement}
\subsection{Analytical definitions}
\textit{Counts and denominators.} We report both absolute paper counts and label shares: counts describe the scale of identified evaluation activity, whereas shares describe its composition. For each label, the denominator is the number of included records with valid coding for that field. Cross-field statistics require all relevant fields to be valid. Multi-label percentages need not sum to 100\%, and domain groups overlap. Unclear and unreported classifications remain in the denominators, as do applicable language labels for nonlinguistic tasks or multilingual support without a complete language list. Additional summaries exclude records carrying these labels. An unselected label is not proof that a property is absent from the real resource.

\textit{Time comparisons.} Two series are presented together: complete calendar years from 2022 through 2025, and January--August in each year from 2022 through 2026. The first describes full-year developments; the second includes 2026 with the same months observed in every year. To assess the effect of choosing one calendar window rather than the other, we hold the start and end years constant. Comparing a series ending in 2025 with one ending in 2026 also changes the endpoint. The overlapping series are not independent replications.

Detailed decomposition, version-year, and bounded-error comparisons use the larger January--August cohorts of 2024 and 2026; full-year and adjacent-year decompositions provide additional context. Early cohorts remain visible, but small cells, especially groups with fewer than 20 records, are not emphasized as trend evidence. Dates refer to first submission, while labels describe the version read. Thus, the comparisons neither track changes in the same resources nor reconstruct their designs at initial release. Additional checks examine results by the analyzed version's year, including records whose first submission and analyzed version fall in the same year, and subsets with explicitly specified task languages.

\textit{Composition and within-group change.} An overall label share can change as the mix of target systems changes, even if each group's label share remains constant. The share can also change within groups. We separate these contributions using Kitagawa's symmetric two-component decomposition \cite{kitagawa1955components}. Let $R_t$ be the overall label share, $r_{tg}$ its share within group $g$, and $w_{tg}$ that group's fraction of the analyzed records in period $t$. The difference between periods 0 and 1 is
\begin{equation}
R_1-R_0=\sum_g(w_{1g}-w_{0g})\frac{r_{1g}+r_{0g}}{2}
+\sum_g(r_{1g}-r_{0g})\frac{w_{1g}+w_{0g}}{2}.
\end{equation}
The first term captures changing group proportions; the second captures changing label shares within groups. These are arithmetic contributions under the chosen grouping, not causal effects. Agent/non-agent is the primary grouping; finer groupings provide sensitivity checks. Early agent cells contain only four records in full-year 2022 and one in January--August 2022, so their decomposition is not used as principal evidence.

\textit{Material generation and scoring.} We ask whether generative-model material and LLM judging occur together more often than their separate frequencies would suggest. We compare the observed joint share with the product of the two individual label shares, the joint share expected under independence. Their difference describes the excess co-occurrence; their ratio (lift) expresses co-occurrence relative to that baseline.

\textit{Sensitivity to hypothetical errors.} We ask how much the observed differences could change under specified error budgets; these scenarios do not estimate actual error rates. For a binary label with $k$ positives among $n$ valid records in a year, allow at most $b=\lfloor en\rfloor$ labels to be flipped, where $e$ is the assumed maximum error fraction. Its rate can then lie between $\max(0,k-b)/n$ and $\min(n,k+b)/n$. Cross-year change bounds combine opposite endpoints, allowing errors to act in the most adverse directions without assuming independence. Budgets of 1\%, 3\%, 5\%, and 10\% are hypothetical per-label, per-year limits, not confidence intervals or measured error rates. A separate scenario removes up to the same fraction of potentially ineligible included records. Neither exercise covers upstream omissions, group misclassification, family duplication, or their joint effects.

All cross-analyses are exploratory and were developed after inspecting data; results were not selected by statistical significance.

\subsection{Flow accounting and field availability}\label{app:availability}
The metadata frame contains 1,153,355 records. M0 routes 173,032 to abstract screening, which excludes 131,525 and does not route 25,079 uncertain records. The resulting 16,428 full-text candidates include 52 unavailable sources. Of 16,376 processed sources, final coding excludes 1,424 and leaves 185 unresolved. The main analysis includes \datasetN\ records: 14,736 with all fields valid and 31 with at least one missing or invalid field (Fig.~\ref{fig:flow}).

Table~\ref{tab:years} shows the annual coverage. Matched January--August cohorts contain 69, 376, 1,683, 3,284, and 5,734 records from 2022 through 2026. Target-system, domain, and modality fields pass the output checks for all included records. Evaluation setup, material origin, scoring source, and language are missing or invalid for 1, 1, 3, and 26 records, respectively.

Among valid outputs, 35 material-origin records carry an unclear or unreported classification, and 17 scoring-source records carry an unclear classification. Another 304 language records carry an unclear, nonlinguistic, or incompletely specified multilingual label. The \texttt{other} domain label appears in 566 records. These permitted labels remain separate from missing or invalid outputs.

\subsection{Complete temporal and domain comparisons}\label{app:windows}
Table~\ref{tab:windows} presents the two temporal series. Agent/tool-system and LLM-scoring shares rise in both series. Interaction is also more frequent in later cohorts, although its small early January--August cohorts do not form a monotonic sequence. Fixed collections remain common throughout. The zero LLM-scoring count in 2022 is an observation about this selected corpus, not evidence that such assessment was absent from the entire field.

The longer view adds an important distinction for material generation. Its full-year share rises from 15.6\% in 2022 to 37.6\% in 2023 and 49.5\% in 2024, followed by 52.5\% in 2025. The matched series likewise rises early but fluctuates around one half in 2024--2026. Thus, an early expansion of model involvement in material creation coexists with a much less pronounced recent change. Figures~\ref{fig:targets}--\ref{fig:joint} provide more detailed views of the larger 2024--2026 matched cohorts.

\input{temporal_table}
\input{domain_annual}

Falling shares do not necessarily mean fewer papers. Language/communication records increase from 334 to 647 and mathematics/formal reasoning from 210 to 528 between the 2024 and 2026 matched cohorts despite declining shares, from 19.8\% to 11.3\% and 12.5\% to 9.2\%, respectively. Social sciences/humanities grows from 191 to 640 while retaining approximately an 11\% share. These domains grow in absolute numbers, but at different rates relative to the identified corpus.

Longer trajectories also resist a single account of increasing specialization. General/cross-domain coding first increases and then decreases in both temporal series; its 2024--2026 matched decline from 47.1\% to 39.0\% is not a continuous decline since 2022. It may coexist with specific domains, so its complement is not a clean measure of exclusively specialized evaluation. Software/coding also fluctuates in the early years rather than rising continuously.

\FloatBarrier
\subsection{Additional modality, material, and language results}\label{app:materials}
The modalities of inputs and scored outputs show a related expansion beyond text, which remains nearly universal. Between the 2024 and 2026 matched cohorts, image coding changes from 24.7\% to 26.1\%, code from 15.6\% to 19.0\%, structured data from 11.3\% to 21.1\%, and audio from 2.0\% to 5.9\%. Video reaches 8.3\% in 2025 and 8.0\% in 2026, illustrating that not all categories increase monotonically. A resource can involve several modalities (Fig.~\ref{fig:materials}).

\textit{Material sources.} Generative-model material appears in 49.5\%, 53.1\%, and 51.3\% of the January--August cohorts in 2024, 2025, and 2026, respectively. Following its earlier expansion (Table~\ref{tab:windows}), its share fluctuates around one half rather than continuing to rise in each recent year. Between 2024 and 2026, program/simulation material increases from 25.0\% to 32.7\%, while human/real-world material changes from 86.3\% to 79.0\%. Sources may coexist; these shares do not show complete replacement of human or real-world materials, and they do not measure human work in design or review.

\textit{Task languages.} English-only coding remains common, with no consistent annual increase across both time series. It changes from 1,332/1,677 (79.4\%) in January--August 2024 to 78.5\% in 2025 and 4,723/5,728 (82.5\%) in 2026. Complete-year shares from 2022 through 2025 are comparatively stable at 77.5\%, 78.2\%, 79.2\%, and 78.6\%, respectively. Here ``English-only'' means that only English was identified in the coded language list, not that other languages were independently ruled out. Incomplete language lists and unsupported language codes, together with the window-dependent size of the change, make this a background description rather than strong evidence of increasing monolinguality.

\subsection{Alternative scoring comparisons}\label{app:scoring}
Additional comparisons show a similar predominance of within-group change. The complete-year 2024--2025 increase is 7.96 points, allocated to 0.05 composition and 7.91 within-group points. This comparison has different endpoints from the matched-cohort comparison and does not isolate calendar-window effects. Restricting the 2024--2026 matched comparison to records whose first submission and analyzed version fall in the same year gives an aggregate increase of 14.64 points and a within-group contribution of 13.84 points. Finer target-system partitions also retain a large within-group contribution.

The pattern is not universal across domains. Between January--August 2024 and 2026, engineering's LLM-scoring share changes from 20/61 (32.8\%) to 155/509 (30.5\%), while the other 12 observed domain groups increase.

\subsection{\texorpdfstring{Generation--scoring association}{Generation-scoring association}}\label{app:association}
The joint share in January--August 2024, 2025, and 2026 is 283/1,681 (16.8\%), 764/3,282 (23.3\%), and 1,450/5,734 (25.3\%), respectively. Co-occurrence becomes more frequent, but lift decreases from 1.32 in 2024 to 1.22 in 2026. Most of the joint-share increase accompanies a rise in the independent baseline, the product of the two individual label shares. The labels do not identify the same model, component, or item across roles, and thus do not establish a self-evaluation loop.

\subsection{Sensitivity to assumed errors}\label{app:sensitivity}
Four larger increases remain positive if at most 5\% of records in each of the 2024 and 2026 January--August cohorts have the particular binary label flipped (Table~\ref{tab:error}). For English-only coding and the smaller material/fixed-collection differences, the bounds allow either an increase or a decrease. Among the 13 domain categories, only the general/cross-domain and language/communication decreases remain negative under a 3\% per-year error budget. These results favor emphasizing larger changes and describing smaller ones cautiously; they do not show that the assumed budgets hold in practice.

\begin{table}[tb]\centering\small
\caption{Conditional sensitivity to binary-label errors in each endpoint cohort. Units are percentage points, 2026 minus 2024, January--August. The budget is assumed, not measured.}\label{tab:error}
\begin{tabular}{lrr}\toprule
Label & Observed change & Adverse range at 5\% per year\\\midrule
Agent/tool system &+19.29&[+9.31, +29.27]\\
Interaction &+18.66&[+8.68, +28.64]\\
LLM scoring &+14.51&[+4.53, +24.49]\\
Execution/environment scoring &+11.55&[+1.56, +21.53]\\
Fixed collection &$-0.75$&[$-10.01$, +9.23]\\
Generative-model material &+1.81&[$-8.17$, +11.79]\\
English-only coding &+3.03&[$-6.92$, +12.97]\\\bottomrule
\end{tabular}
\end{table}

Separately allowing removal of up to 5\% of potentially ineligible records per year gives change lower bounds of +15.05, +14.39, +10.02, and +6.90 points for agent, interaction, LLM-scoring, and execution-scoring labels. This deletion-only scenario is not combined with label flips and does not account for missed papers.

The earlier targeted interaction review documented two omissions in the 2026 cohort. Adding these missing labels would change that cohort's share by approximately 0.035 percentage points, with little effect on the 2024--2026 comparison. This is not an adjustment for all known or unobserved errors.

%% file: temporal_table.tex
\begin{table}[tb]\centering\small
\caption{Two temporal series of included papers and selected design shares (percentages). Shares use field-valid denominators; labels overlap. Gen. material denotes generative-model participation in evaluation materials. The two panels overlap and are not independent replications.}\label{tab:windows}
\begin{tabular}{lrrrrrr}\toprule
Year & Papers & Agent & Interaction & Gen. material & LLM judge & Fixed\\\midrule
\multicolumn{7}{l}{\textit{Full calendar years}}\\
2022 & 160 & 2.5 & 5.6 & 15.6 & 0.0 & 93.8\\
2023 & 861 & 7.4 & 9.9 & 37.6 & 18.0 & 96.3\\
2024 & 2,735 & 9.5 & 10.8 & 49.5 & 28.0 & 95.3\\
2025 & 5,277 & 15.2 & 16.7 & 52.5 & 35.9 & 95.1\\
\midrule
\multicolumn{7}{l}{\textit{January--August in each year}}\\
2022 & 69 & 1.4 & 8.7 & 14.5 & 0.0 & 92.8\\
2023 & 376 & 6.1 & 8.2 & 28.5 & 13.6 & 96.3\\
2024 & 1,683 & 9.0 & 9.5 & 49.5 & 25.8 & 95.7\\
2025 & 3,284 & 13.0 & 15.4 & 53.1 & 35.5 & 95.3\\
2026 & 5,734 & 28.3 & 28.2 & 51.3 & 40.3 & 95.0\\
\bottomrule
\end{tabular}
\end{table}

%% file: domain_annual.tex
\begin{table}[tb]\centering\small
\caption{Professional and application domains: Complete calendar years, 2022--2025. Cells show paper counts with within-year percentages in parentheses. Domain coding is valid for all included records; categories overlap. Counts are identified papers, not independent families or adoption measures.}\label{tab:domains-annual}
\setlength{\tabcolsep}{3pt}
\begin{tabular}{lrrrr}\toprule
Domain & 2022 & 2023 & 2024 & 2025\\\midrule
Healthcare/biomedicine & 9 (5.6) & 64 (7.4) & 281 (10.3) & 585 (11.1)\\
Finance/business & 5 (3.1) & 47 (5.5) & 172 (6.3) & 434 (8.2)\\
Engineering & 2 (1.2) & 46 (5.3) & 120 (4.4) & 368 (7.0)\\
Cybersecurity & 1 (0.6) & 28 (3.3) & 88 (3.2) & 235 (4.5)\\
Software/coding & 28 (17.5) & 132 (15.3) & 456 (16.7) & 909 (17.2)\\
Natural sciences & 7 (4.4) & 63 (7.3) & 198 (7.2) & 470 (8.9)\\
Law/public policy & 4 (2.5) & 22 (2.6) & 84 (3.1) & 182 (3.4)\\
Education & 0 (0.0) & 14 (1.6) & 49 (1.8) & 109 (2.1)\\
Social sci./humanities & 15 (9.4) & 109 (12.7) & 313 (11.4) & 622 (11.8)\\
Language/communication & 90 (56.2) & 237 (27.5) & 520 (19.0) & 787 (14.9)\\
Math./formal reasoning & 16 (10.0) & 106 (12.3) & 323 (11.8) & 614 (11.6)\\
General/cross-domain & 53 (33.1) & 420 (48.8) & 1,275 (46.6) & 2,215 (42.0)\\
Other & 5 (3.1) & 15 (1.7) & 79 (2.9) & 213 (4.0)\\
\midrule
All included papers & 160 & 861 & 2,735 & 5,277\\\bottomrule
\end{tabular}
\end{table}